\documentclass{article}

\usepackage{microtype}
\usepackage{graphicx}
\usepackage{subfigure}
\usepackage[justification=centering]{caption}
\usepackage{booktabs} 

\usepackage{amsmath,amssymb}
\DeclareMathOperator{\E}{\mathbb{E}}
\usepackage{bm}

\usepackage{hyperref}

\usepackage[algo2e]{algorithm2e} 

\usepackage[accepted]{icml2020}

\DeclareMathOperator{\diag}{diag}
\SetCommentSty{mycommfont}

\begin{document}

\twocolumn[
\icmltitle{Combining Reinforcement Learning  and Inverse  Reinforcement Learning for 
Asset Allocation Recommendations
}

\begin{icmlauthorlist}
	\icmlauthor{Igor Halperin}{fid}
	\icmlauthor{Jiayu Liu}{fid}
	\icmlauthor{Xiao Zhang}{fid}

\end{icmlauthorlist}

\icmlaffiliation{fid}{AI Center of Excellence for Asset Management, Fidelity Investments}
\icmlcorrespondingauthor{Igor Halperin, JiayuLiu, Xiao Zhang}{igor.halperin@fmr.com, jiayu.liu@fmr.com, xiao.zhang2@fmr.com}
\vskip 0.3in
]

\printAffiliationsAndNotice{} 

\begin{abstract}
We suggest a simple practical method to combine 
the human and artificial intelligence  to both learn best investment practices of fund managers, and provide recommendations to improve them.   
Our approach is based on a combination of  Inverse Reinforcement Learning (IRL) and RL. 
First, the IRL component learns the intent of fund managers as suggested by their trading history, and recovers their implied reward function. At the second step, this reward function is used by a direct RL algorithm to optimize asset allocation decisions. We show that our method is able to improve over the performance of individual fund managers.
   
\end{abstract}

\section{Introduction}

Portfolio management is a quintessential example of stochastic multi-period (dynamic) optimization, also known as stochastic optimal control. Reinforcement Learning (RL) provides data-driven methods for problems of optimal control, that are able to work with high dimensional data.
Such applications are beyond the reach of classical methods of optimal control, which  typically only  work for low-dimensional data. Many of the most classical problems of quantitative finance such as portfolio optimization, wealth management and option pricing can be very naturally formulated and solved as RL problems, see e.g. \cite{MLF}. 

A critically important input  to any RL method is a reward function. The reward function provides a condensed formulation of the agent's \emph{intent}, and respectively it informs the optimization process of RL. The latter amounts to finding policies that maximize the expected total reward obtained in a course of actions by the agent. One simple example of a reward function for RL of portfolio management is provided by the classical Markowitz mean-variance objective function applied in a multi-period setting (\cite{MLF}). However, there might exist other specifications of the reward function for portfolio management that might be more aligned with the actual practices of active portfolio managers.

Inverse Reinforcement Learning (IRL) is a sub-field of 
RL that addresses the inverse problem of inferring the reward function from a demonstrated behavior. The behavioral data are given in the form of sequences of state-action pairs performed either by a human or AI agent. IRL is popular in robotics and video games for cases where engineering a good reward function (one that promotes a desired behavior) may be a challenging task on its own. The problem of dynamic portfolio optimization arguably belongs in such class of problems. Indeed,  while the 'best' objective function to be used for portfolio optimization is unknown, we can try to use IRL to \emph{learn} it from the investment behavior of active portfolio managers (PMs). 

In this paper, we propose a combined use of IRL and RL (in that sequence) to both learn the investment strategies of portfolio managers, and \emph{refine} them, by suggesting tweaks that improve the overall portfolio performance. Our approach can be viewed as a way to develop a `collective intelligence' for a group of fund managers with similar investment 
philosophies. By using trading histories of such funds, our algorithm learns the reward function from all of them jointly in the IRL step, and then uses this reward function to optimize the investment policy in the second, RL step.

We emphasize that our approach is not intented to \emph{replace} active portfolio managers, but rather to 
\emph{assist} them in their decision making. In particular, we do not re-optimize the stock selection made by PMs, but simply change allocations to stocks that were already selected. As will be explained in more details below, data limitations force us to work with a dimensionally reduced portfolio optimization problem. In this work, we choose to aggregate all stocks in a given portfolio by their industry sector. This provides a low-dimensional view of portfolios held by active managers. Respectively, the optimal policy is given in terms of allocations to industrial sectors, and not in terms of allocations to individual stocks. However, given any portfoflio-specific selection of stocks for a given sector, a recommended optimal sector-level allocation can be achieved simply by changing positions in stocks already selected.\footnote{This assumes that stocks from all sectors are already present in the portfolio, and that changes of  positions in individual stocks do not produce a considerable price impact.}  
      
Our paper presents a first practical example of a joined application of IRL and RL for asset  allocation and portfolio management that leverages human intelligence and transfers it, in a digestible form, to an AI system (our RL agent). While we use specific algorithms for the IRL and RL steps (respectively, the T-REX and G-learner algorithms, see below), our framework is general and modular, and enables using of other IRL and RL algorithms, if needed.
We show that a combined use of  IRL and RL enables to both learn best investment practices of fund managers, and  provide recommendations to improve them.  

The paper is organized as follows. In Sect.~\ref{sect_Related_work} we overview previous related work. Sect.~\ref{sect_Method} outlines the theoretical framework of our approach, presenting the parametric T-REX algorithm for the IRL part, and the G-learner algorithm for the RL part. Sect.~\ref{sect_Experiments} shows the results of our analysis. The final Sect.~\ref{sect_Summary} provides a summary and conclusions. 


\subsection{Related Work}
\label{sect_Related_work}

Our approach combines approaches developed separately in the research community focused on RL and IRL problems.
For a general review of RL and IRL along with  their applications to finance, see \cite{MLF}.  

In its reliance on a combination  of IRL and RL for  optimization of a financial portfolio, this   
paper follows the  approach of   \cite{dixon2020g}. The latter uses a continuous state-action version of G-learning,  
a probabilistic extension of Q-learning 
 \cite{glearning}, for the RL step, and an IRL version of G-learning called GIRL for the IRL step. In this work,  we  retain the  G-learner  algorithm from \cite{dixon2020g} for the RL step, but replace the  IRL step  with  a  
parametric version of a recent algorithm called T-REX
\cite{brown2019extrapolating}  which will be presented in more details in Sect.~\ref{trex}.

\section{The method}
\label{sect_Method}

In this section, we describe our proposed framework for asset allocation recommendations as a sequence that combines IRL and RL in a two-step procedure. The framework is designed to be generic enough to accommodate various IRL and RL methods. In this work, we focus on an extension of the approach of \cite{dixon2020g} that incorporates portfolio performance ranking information into its policy optimization algorithm. This extension is inspired by a recently proposed IRL method called T-REX \cite{brown2019extrapolating}. 
Here we first provide a short overview of T-REX and G-learner methods, and then introduce our IRL-RL framework. In its IRL step, our proposed parametric T-REX algorithm infers PMs'  asset allocation decision rules from portfolio historical data, and provides a high quality reward function to evaluate the performance. The learned reward function is then utilized by G-learner in the RL step to learn the optimal policy for asset allocation recommendations. 
Policy optimization amounts to a recursive  procedure involving only linear algebra operations, which can be computed within  seconds. 

\subsection{The IRL step with parametric T-REX}
\label{trex}

IRL deals with recovering reward functions from the observed agents' behavior as a way of rationalizing their intent. The learned reward function can then be used for evaluation and optimization of action policies. The IRL approach reduces the labor of defining proper reward functions in RL for applications where their choice  is not immediately obvious.  
 
Many traditional IRL methods try to learn the reward  function from demonstrations under the assumption that the demonstrated history corresponds to an optimal or near-optimal policy employed by an agent that produced the data, see e.g. \cite{pomerleau1991efficient,ross2011reduction,ijcai2018-687}. 
These IRL methods therefore simply try to mimic or justify the observed behavior by 
finding a proper reward function that explains this behavior. This implies that by following their approach, the best one can hope for IRL for investment decisions 
is that it would be able to \emph{imitate} investment policies of asset managers.

However, the assumption that the demonstrated behavior \emph{is} already optimal might not be realistic, or even verifiable, for applications in quantitative finance. This is because, unlike robotics or video games, the real market environment cannot be reproduced on demand.  
Furthermore, relying on this assumption for inferring rewards may also have a side effect of transferring various psychological biases of asset managers into an inferred `optimal' policy. 

Instead of simply trying to mimic the observed behavior of asset managers, it would be highly desirable to try to \emph{improve} over their investment decisions, and potentially correct for some biases implicit in their decision-making. 
A principled way towards implementing such a program would be to infer 
the \emph{intent} of asset managers from observing their trading decisions  \emph{without} the assumption they their strategies were fully or partially successful. With such an approach, demonstrated trajectories would be scored by how well they succeed in achieving a certain goal.
This is clearly very different from seeking a reward function that would simply rationalize the observed behavior of asset managers.  

Such inference of agents' intent is made possible using
a recently proposed IRL algorithm called T-REX, for Trajectory-based Reward Extrapolation \cite{brown2019extrapolating}. 
The T-REX method finds the reward function that captures the \emph{intent} of the agent \emph{without} assuming a  near-optimality of the demonstrated behavior. Once the reward function that captures agent's intent is found, it can be used to further improve the performance by optimizing this reward. This could be viewed as extrapolation of the reward beyond the demonstrated behavior, thus explaining the `EX' part in the name T-REX.

To overcome the issue of potential sub-optimality of demonstrated behavior, T-REX relies on certain extra information that is not normally used in other IRL methods.
More specifically,  T-REX uses rankings of all demonstrated trajectories according to a certain final score. The final score is assigned to the whole trajectory rather than to a single time-step. The local reward function is then learned from the condition that it should promote the most desirable trajectories according to the score. In this work, we   score different portfolio trajectories by their total realized returns,
however, we could also use risk-adjusted metrics such as the 
Sharpe or  Sortino ratios.
  

Let $\bm{O}: \{\bm{S}, \bm{A}\} $ be a state-action space of an MDP environment, and $\hat{r}_{\bm{\theta}}(\cdot)$ with parameters $\bm{\theta}$ be a target reward function to be optimized in the IRL problem. Given $M$ ranked observed sequences $\{\bm{o_m}\}_{m=1}^{M}$ ($\bm{o_i} \prec \bm{o_j}$ if $i < j$, where ``$\prec$" indicates the preferences,  expressed e.g.  via a ranking order, between pairwise sequences), T-REX conducts reward inference by solving the following optimization problem: 
\begin{equation}
\label{trex_opt}
\max_{\bm{\theta}} \sum_{\bm{o_i} \prec \bm{o_j}}{\log \frac{e^ {\sum\nolimits_{\{\bm{s},\bm{a}\} \in \bm{o_j}}{\hat{r}_{\bm{\theta}}(\bm{s},\bm{a})}}}{e^{\sum\nolimits_{\{\bm{s},\bm{a}\} \in \bm{o_i}}{\hat{r}_{\bm{\theta}}(\bm{s},\bm{a})}} + e^{\sum\nolimits_{\{\bm{s},\bm{a}\} \in \bm{o_j}}{\hat{r}_{\bm{\theta}}(\bm{s},\bm{a})}}}}
\end{equation}
This objective function is equivalent to the softmax normalized cross-entropy loss for a binary classifier, and can be easily trained using common machine learning libraries such as PyTorch or TensorFlow.  As a result, the learned optimal reward function can preserve the ranking order between pairs of sequences. 

The original T-REX algorithm is essentially a non-parametric model that encodes the reward function into a  deep neural network (DNN). This might be a reasonable approach for robotics or video games where there is a plenty of data, while the reward function might be highly non-linear and complex. Such setting is however ill-suited for applications to portfolio management where the amount of available data is typically quite small.
Therefore, in this work we pursue a parametric version 
of T-REX, where the reward function is encoded into  a function with a small number of parameters, which is then 
optimized using Eq.(\ref{trex_opt}).  A particular model 
of the reward function that will be presented in the next section has only 4 tunable parameters, which appears to be about the right order of model complexity given available data. The added benefit in comparison to the original DNN-based implementation of T-REX is that a parametric T-REX is much faster to train.



\subsection{The RL step with G-learner}
\label{glearner}

G-learner is an algorithm for the direct RL problem of portfolio optimizaiton that was proposed in \cite{dixon2020g}. It relies on 
a continuous state-action version of G-learning, a probabilistic extension of Q-learning designed for noisy environments \cite{glearning}.
G-learner solves a finite-horizon direct RL problem of portfolio optimization in a high-dimensional continuous state-action space using a sample-based approach that operates with available historical data of portfolio trading.  
The algorithm uses a closed-form state transition model to capture the stochastic dynamics in a noisy environment, and therefore belongs in the class of model-based RL approaches. In this work, we 
employ G-learner as a RL solver for the sequential portfolio optimization, while defining 
a new reward function to evaluate portfolio managers' performance.     

Unlike \cite{dixon2020g} that considered G-learning for a portfolio of individual stocks  held by a retail investor, here we consider portfolios typically held by professional asset managers.
Due to data limitations, it is unfeasible to estimate models
that  track the whole investment universe  counting thousands of stocks, and
maintain individual stocks holdings and their changes as, respectively, state and action variables. 
A viable alternative is is to aggregate individual stock 
holdings into buckets constructed according to a particular dimension reduction principle, and define the state and action variables directly in   such dimensionally reduced space. In this paper, we choose to aggregate all stocks in the PM's portfolio into $ N = 11$ sectors of the standard GICS industry classification, effectively mapping the portfolio into portfolio of sector exposures.  The  state vector  $\bm{x_t} \in R^N$ is respectively defined as a vector of  dollar values of stock positions in each sector at time $t$.  The action variable  $\bm{u_t} \in R^N$ is 
given by the vector of changes in these positions as a result of trading at time step $ t $. 
Furthermore, let $\bm{r_t} \in R^N$ represent the asset returns as a random variable with the mean $\bm{\bar{r}_t}$   and covariance matrix  $\bm{\Sigma_r}$. As in  our approach  we identify assets with sector exposures, $\bm{\bar{r}_t}$   and $\bm{\Sigma_r}$ will be interpreted as expected sector returns and sector return covariance, respectively.

The state transition model is defined as follows: 
\begin{equation}
\label{transition}
\bm{x_{t+1}} = \bm{A_t} (\bm{x_t} + \bm{u_t}), \bm{A_t} = \diag(\bm{1} + \bm{r_t}) 
\end{equation} 

In this work we use a simplified form of the reward from \cite{dixon2020g}, which we  write as follows: 
\begin{equation}
\label{reward_func}
\begin{split}
R_t(\bm{x_t}, \bm{u_t}|\bm{\theta}) = & - \E_t\left[\left(\hat{P}_{t} - V_{t}\right)^2\right] \\  
& -\lambda \cdot \left(\bm{1}^T \bm{u_t} -C_t\right)^2 - \omega \cdot \bm{u_t}^T \bm{u_t} \\
\end{split}
\end{equation} 
where $ C_t $ is  a money flow  to the fund at time step $ t $ minus the change of a cash position of the fund at the same time step, $ \lambda $ and 
$ \omega $ are parameters, and $ V_t $  and  $ \hat{P}_t $ are, respectively, the values of asset manager's and reference portfolios, defined as follows:
\begin{equation}
\label{term1}
\begin{split}
V_{t} & = (\bm{1} + \bm{r_t} )^T(\bm{x_t} + \bm{u_t}) \\
\hat{P}_{t} & = \rho \cdot B_t + (1-\rho) \cdot \eta \cdot \bm{1}^T \bm{x_t} 
\end{split}
\end{equation}
where $ \eta $ and $ \rho $ are additional parameters, 
and $ B_t $ is a benchmark portfolio, such as e.g. the SPX index portfolio,  properly re-scaled to match the size of the PM's portfolio at the start of the investment period. 

The reward function (\ref{reward_func}) consists of three terms, each encoding one of the portfolio managers' trading insights.  In the first term, $\hat{P}_{t}$ defines the target portfolio market value at time $t$. It is specified as a linear combination of a reference benchmark portfolio value $B_t$ and the current portfolio growing  
with rate $\eta$ according to Eq.(\ref{term1}), where $\rho \in [0,1]$ is a parameter defining the relative weight between the two terms. $V_{t} $ gives the portfolio value at time 
$ t + \Delta t $, after the trade $\bm{u_t}$ is made at time $ t $. The first term in (\ref{reward_func})  imposes a penalty for under-performance of the traded portfolio relative to its moving target. The second term enforces the constraint 
that the total amount of trades in the portfolio should match
the inflow $ C_t $ to the portfolio at each time step, with $\lambda$ being a parameter penalizing violations of the equality constraint. The third term approximates transaction costs  by a quadratic function with parameter $\omega$, thus serving as a $L_2$ regularization. The vector $\bm{\theta}$ of model parameters thus contains four reward parameters $\{\rho, \eta, \lambda, \omega\}$.

Importantly, the reward  function (\ref{reward_func}) is a quadratic function of the state and action. Such a choice of the reward function implies  that  the value-  and action-value functions are similarly quadratic functions of the state and action, with time-dependent coefficients. Respectively, with a quadratic reward functions such as (\ref{reward_func}), a G-learning algorithm should  not engage neural networks or other function approximations, which would  invariably bring their own parameters to be optimized. In contrast,  with the  quadratic reward (\ref{reward_func}),  learning the optimal  value and  action-value functions amounts to learning the coefficients of quadratic  functions, which amounts to a small number of linear algebra operations \cite{MLF}. As a result, an adaptation of the general G-learning method for the quadratic reward  (\ref{reward_func}) produces a very fast policy optimization algorithm called G-learner in \cite{dixon2020g}.      
 
G-learner is applied to solve the proposed RL problem with our defined MDP model in Eq.(\ref{transition}) and reward function in Eq.(\ref{reward_func}). Algorithm \ref{glearner_algo} demonstrates the entire optimization solving process. It takes the model parameters as inputs, along with  discount factor $\gamma$. In addition, it uses 
a prior policy  $\bm{\pi}^{(0)}$ which usually encodes domain knowledge of real world problems. G-learner (and G-learning in general) control the deviation of the optimal policy 
$\bm{\pi_t}$ from the prior  $\bm{\pi}^{(0)}$ 
by incorporating the KL divergence of $\bm{\pi_t}$ and 
$\bm{\pi}^{(0)}$ into a modified, regularized reward function, with 
a hyperparameter $ \beta $ that controls the magnitude of the KL regularizer.
When $\beta $ is large, the deviation can be arbitrarily large, while in the limit $ \beta \rightarrow 0 $, $\bm{\pi_t}$ is forced to be equal to $\bm{\pi}^{(0)}$, so there is no learning in this limit.
Furthermore, 
$\bm{F_t} (\bm{x_t}) $ is the value function, and $\bm{G_t} (\bm{x_t}, \bm{u_t})$ is the action-value function. 
The algorithm is initialized by computing the optimal terminal value and action-value functions $\bm{F_T}^*$ and $\bm{G_T}^*$.
They are computed by finding the optimal last-step action $ \bm{a}_T $ that achieves the highest one-step terminal reward, i.e. solves the equation $ \frac{\partial{R_t(\bm{x_t}, \bm{u_t})}}{\partial{\bm{u_t}}} |_{ t=T} = 0$.
Thereafter, the policy associated with earlier time steps can be derived recursively backward in time as shown in the {\it{for}} loop of Algorithm \ref{glearner_algo}. For a detailed derivation of $Value\_Update$ and $ActionValue\_Update$ steps, we refer to Section 4 in \cite{dixon2020g}. 

\begin{algorithm}
	\SetKwFunction{isOddNumber}{isOddNumber}
	\SetKwFunction{VFU}{Value\_Update}
	\SetKwFunction{VAFU}{ActionValue\_Update}
	\SetKwInOut{KwIn}{Input}
	\SetKwInOut{KwOut}{Output}
	
	\KwIn{$ \lambda, \omega, \eta, \rho, \beta, \gamma, \{\bm{\bar{r}_t}, \bm{x_t}, \bm{u_t}, B_t, C_t\}_{t=0}^{T}, \bm{\Sigma_r},  \bm{\pi}^{(0)}$}
	\KwOut{$ \bm{\pi_t}^{\ast} = \bm{\pi}^{(0)} \cdot e^{\beta \left(\bm{G_t}^{\ast} - \bm{F_t}^{\ast} \right)}, t = 0, \cdots, T$} 
	
	\textbf{Initialize: } $\bm{F_T}^{\ast}, \bm{G_T}^{\ast}$ 
	
	\While {not converge}  { 
		\For {t $\in$ [T-1, -1, 0]} {    
		$\bm{F_t} \gets \VFU(\bm{F_{t+1}}, \bm{G_{t+1}})$ \\
		$\bm{G_t} \gets \VAFU(\bm{F_t}, \bm{G_{t+1}})$
	    }
    }

	\KwRet{$\{\bm{F_t}^{\ast}, \bm{G_t}^{\ast},   \bm{\pi_t}^{\ast} \}_{t=0}^{T}$}
	\caption{G-learner policy optimization}
	\label{glearner_algo}
\end{algorithm}

Once the optimal policy $\bm{\pi_t}^* = \bm{\pi_t}^*(\bm{u_t}|\bm{x_t})$ is learned,  the recommended action at time 
$t$ is given by the mode of the action policy for the given state $  \bm{x_t}$.
 
\subsection{A unified IRL-RL framework}
\begin{figure}[ht!]
	\centering
	\includegraphics[width=0.75\linewidth]{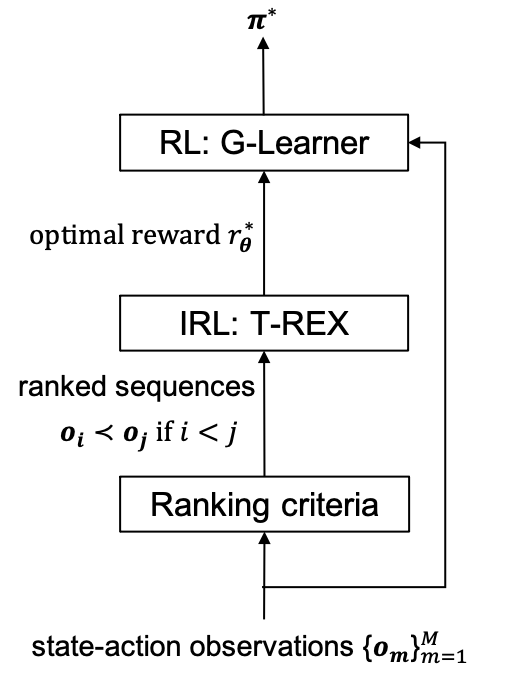}
	\caption{The flowchat of our IRL-RL framework}
	\label{flowchart}
\end{figure}

Figure \ref{flowchart} illustrates the overall workflow of our proposed IRL-RL framework. We apply this framework to provide portfolio allocation recommendations to improve fund returns. It takes the observed $M$ state-action sequences ($\bm{o_m}: \{\bm{x_t}, \bm{u_t}\}_{t=0}^T$) from multiple funds as input training data. As cumulative funds'  returns can be used to measure the  overall PM's performance over a given observation period, we use the realized total returns in the train dataset  
as a ranking criterion for the IRL step with the parametric T-REX. As an alternative, we could rely on risk-adjusted  metrics such as e.g. the Sharpe or Sortino ratios.
 
With the chosen ranking criterion, the T-REX IRL module finds the optimal parameters of the reward function defined in Eq.(\ref{reward_func}).  The optimal reward function parameters are then passed to the RL module which optimizes the action policy. Both modules operate in the offline learning mode, and do not require an online integration with the trading environment.  

The IRL module can also work individually to provide behavioral insights via analysis of inferred model parameters. It is designed to be generalizable to different reward functions, if needed. Ranking criteria need to be rationally determined to align with the logic of rewards and depict the intrinsic behavioral goal of demonstrations. 
The human-provided ranking criterion based on  the performance on different portfolio paths thus informs a pure AI system (our G-learner agent). As a result, our IRL/RL pipeline implements a simple human-machine interaction loop in the context of portfolio management and asset allocation.

We note that by using T-REX with the return-based scoring, our model already has a great starting point to hope for success. Indeed, as our reward function and ranking are aligned with PMs' own assessment of their success, this already defines a `floor' for the model performance. 
This is because our approach is  guaranteed 
to at least match individual funds' performance by taking the exact replica of their actions. Therefore, our method cannot do worse, at least in-sample, than PMs but it can do better - which is indeed the case as will be shown below.    


Another comment due here is that while in this work we use 
the aggregation to the sectors as a particular scheme for dimension reduction, our combined IRL-RL framework could also be used with other ways of dimension reduction, for example we could aggregate all stocks by their exposure to a set of factors.


\section{Experiments}
\label{sect_Experiments}

This section describes our experiments. 
First, we explain data pre-processing steps. Once the state and action variables are computed, we sequentially apply the parametric T-REX and G-Learning algorithms according to  
Eq.(\ref{trex_opt}) and Algorithm \ref{glearner_algo}, respectively. We then show how portfolio managers' insights can be inferred using T-REX, and then demonstrate 
that G-learner is able to outperform active managers by finding optimal policies for rewards inferred at the first stage.

\subsection{Data Preparation}
In our experiments, two sets of mutual funds are chosen based on their benchmark indexes. To anonymize our data, we  replace actual funds' names by single-letter names.
The first set includes six funds ($\{S_i\}_{i=1}^6$) that all have the S\&P 500 index as a benchmark. The second set includes twelve funds 
that have the the Russell 3000 index as a benchmark. 
The second group is further divided into growth and value fund groups ($\{RG_i\}_{i=1}^7$ and $\{RV_i\}_{i=1}^5$). Each fund's trading trajectory covers the period from January 2017 to December 2019, and contains its monthly holdings and trades for eleven sectors (i.e., $\bm{x_t},\bm{u_t}$)  as well as monthly cashflows $C_t$.  The first two years of data are used for training (so that we choose T=24 months), and the last year starting  from  January 2019  is used for testing.  At the starting time step we assign each fund's total net asset value to its corresponding benchmark value (i.e., $B_{t=0}$) in order to align their size and use the actual benchmark return at each time step to calculate their values afterwards (i.e., $t>0$). The resulting time series $\{\bm{x_t}, \bm{u_t}, B_t, C_t\}_{t=0}^{T}$ are further normalized by dividing by their initial values at $t=0$. 
In the training phase, the canonical ARMA model \cite{hannan2009multiple} is applied to forecast expected sector returns $\bm{r_t}$, and the regression residue is then used to estimate the sector return covariance matrix $\bm{\Sigma_r}$. The prior policy $\bm{\pi}^{(0)}$ is fitted to a multivariate Gaussian distribution with a constant mean and variance calculated from sector trades in the training set.

\subsection{Parametric T-REX with portfolio return rankings}
The pre-processed fund trading data is ranked based on funds' overall performance realized over the training period. T-REX is then trained to learn the four parameters  $\{\rho, \eta, \lambda, \omega\}$ that enter Eq.(\ref{reward_func}). The training/test experiments are executed separately for all three fund sets (i.e. funds with the S\&P 500 benchmark, Russell 3000 Growth funds, and Russell 3000 Value funds). 

\begin{figure}[ht!]
	\centering
	\includegraphics[width=0.95\linewidth]{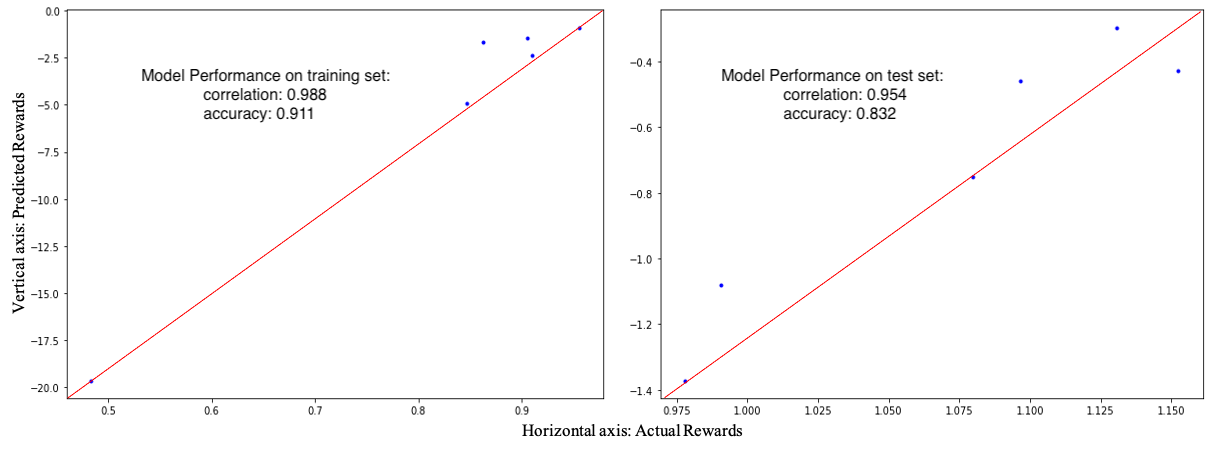}
	\caption{T-REX: classification accuracy and ranking order preservation measured by correlation scores}
	\label{trex_sector_acc}
\end{figure}

Recall that the loss function for T-REX is given by the cross-entropy loss for a binary classifier, 
see Eq.(\ref{trex_opt}). Figure \ref{trex_sector_acc} shows the classification accuracy in the training ($91.1\%$) and test phases ($83.2\%$) for the S\&P 500 fund set. The plots in the two sub-figures show that the learned reward function preserves the fund ranking order. Correlation scores are used to measure their level of association, producing the values of $0.988$ and $0.954$ for the training and test 
sets, respectively. The excellent out-of-sample correlation scores gives a strong support to our chosen method of the 
reward function design.

\begin{figure}[ht!]
	\centering
	\includegraphics[width=0.95\linewidth]{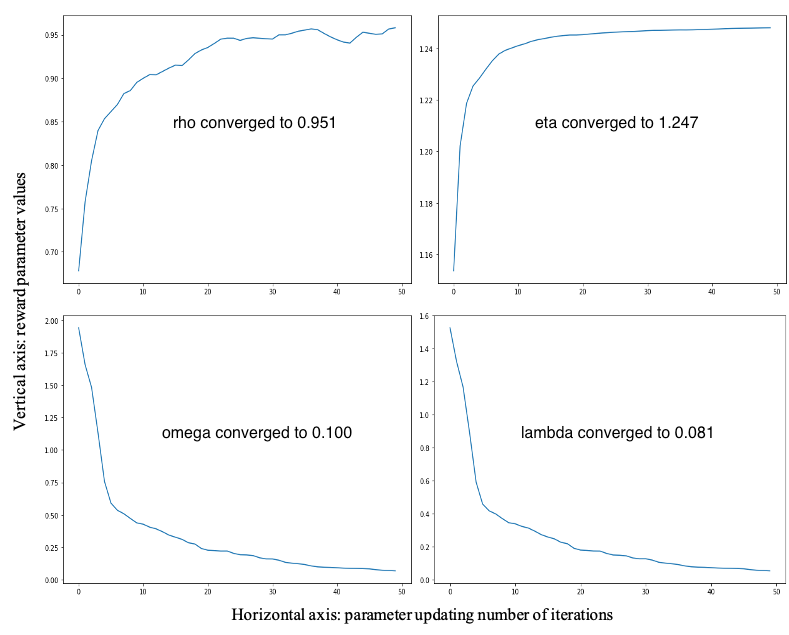}
	\caption{T-REX: reward parameter learning curve}
	\label{trex_sector_param}
\end{figure}

Fund managers' investment strategies can be analyzed using the inferred reward parameters. Figure \ref{trex_sector_param} shows the parameter convergence curve using the S\&P 500 fund group. The values converge after about thirty iterations, which suggests adequacy of our chosen model of the reward to the actual trading data.
The optimized benchmark weight $\rho^{\ast} = 0.951$ captures the correlation between funds' performance and their benchmark index. Its high value implies that portfolio managers' trading strategy is to target an accurate benchmark tracking. The value $\eta^{\ast}=1.247$ implies funds are in the growing trend with prospective rate at $24.7\%$ from fund managers' view. Since the sum of trades are close enough to the fund cashflows in our dataset, the introduced penalty parameter $\lambda$ converges to a very small value at $0.081$. The estimated average trades' volatility $\omega^{\ast}$ is around $10\%$, which measures the level of consistency across trades in the S\&P 500 fund group. 

The experimental results of our proposed parametric T-REX are summarized in Table \ref{trex_param_opt_table}. Our model achieved high classification accuracy ($\mathit{acc}$) and correlation scores ($\mathit{cor}$) on all three fund groups in both training and test phases. The optimal values of reward function parameters are listed in the table as well. Small values of $\lambda^{\ast}$ verify the holding of equality constraint over the sum of trades in our dataset. Comparing across different fund groups, we can identify different trading strategies of portfolio managers. Funds in the S\&P500 group are managed to track their benchmark very closely with correlation $\rho^{\ast} = 0.951 $.  
On the other hand, correlation $\rho^{\ast}  $ reduces significantly to 0.584 and then to 0.186 for fund groups $\{RG_i\}_{i=1}^7$ and $\{RV_i\}_{i=1}^5$ which have the same benchmark index Russell 3000, while being segmented based on their investment approaches (i.e., growth vs. value). For the bucket denoted $\{S_i\}_{i=1}^6$, growth and blend funds are combined. As shown in the table, funds in the segmented groups show very limited variations of their trading amounts compared to those combined in a single group. This can be seen as an evidence of fund managers' consistency in their investment decisions/policies. Growth funds in $\{RG\}_{i=1}^7$ with $\eta^{\ast}=1.532$ produce the highest expected growth rates as implied by their trading activities. This result is in line with the fact that growth investors select companies that offer strong earnings growth while value investors choose stocks that appear to be undervalued in the marketplace.

\begin{table}[ht!]
	\caption{Inferred reward parameter values and T-REX model accuracy from all three fund groups}
	\label{trex_param_opt_table}
	\begin{center}
		\begin{small}
				\begin{tabular}{cccc}
					\toprule
					 Fund Group & $\{S_i\}_{i=1}^6$ & $\{RG_i\}_{i=1}^7$ & $\{RV_i\}_{i=1}^5$ \\
					\midrule
					$\rho^{\ast}$       & 0.951 & 0.186  & 0.584 \\
					$\eta^{\ast}$       & 1.247  & 1.532  &  1.210 \\
					$\lambda^{\ast}$ & 0.081 & 0.009 & 0.009 \\
					$\omega^{\ast}$ & 0.100 & 0.012  & 0.009 \\
					$\mathit{acc}$ (train/test)  & 0.911/0.832   & 0.878/0.796 & 0.906/0.832 \\
					$\mathit{cor}$ (train/test)   & 0.988/0.954 & 0.925/0.884 &  0.759/0.733  \\
					\bottomrule
				\end{tabular}
		\end{small}
	\end{center}
\end{table}
The four reward parameters can be used to group different fund managers or different 
trading strategies into distinct clusters in the space of reward parameters. While the present analysis in this paper suggests pronounced patterns based on the comparison across three fund groups, including more fund trading data from many portfolio managers might be able to bring further insights into fund managers' behavior. This is left here for a future work.

\subsection{G-learner for allocation recommendation}
Once the reward parameters are learned for all groups of funds, they are passed to G-learner for the policy optimization step. 
The pre-processed fund trading data in year 2017 and 2018 is used as inputs for policy optimization (see Figure \ref{flowchart}). Trading trajectories were truncated to twelve months' long in order to align with the length of the test period (year 2019). The optimal policy  $\bm{\pi_t}^* = \bm{\pi_t}^*(\bm{u_t}|\bm{x_t})$ is learned using the train dataset, and then applied to the test set throughout the entire test period. Differently from the training phase, where we use expected sector returns in the reward function, in the test phase we use realized monthly returns to derive the next month's holdings $\bm{x_{t+1}}$ after trades $\bm{u_t}$ was made for the current month with $\bm{x_t}$. At the start time of the test period, the holdings $\bm{x}_{t=0}$ coincide with actual historical holdings. We use this procedure along with the learned policy $\bm{\pi_t}^*(\bm{u_t}|\bm{x_t})$ onward from
February 2019 to create
counterfactual portfolio trajectories for the test period. We will refer to these RL-recommended  trajectories as the AI Alter Ego's (AE) trajectories.

\begin{figure}[ht!]
	\centering
	\includegraphics[width=0.95\linewidth]{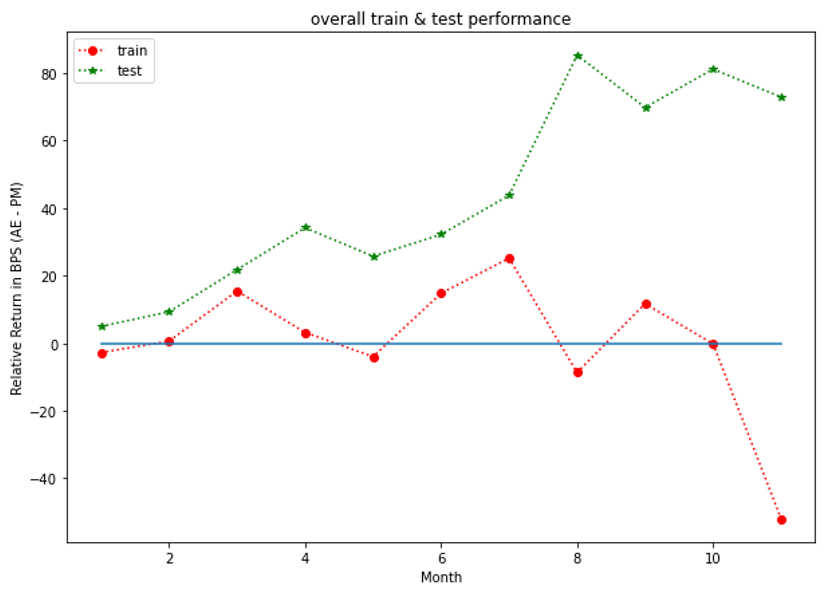}
	\caption{G-learner: overall trading performance with funds benchmarked by S\&P500}
	\label{glearner_acc_avg_sp500}
\end{figure}

\begin{figure}[ht!]
	\centering
	\includegraphics[width=0.95\linewidth]{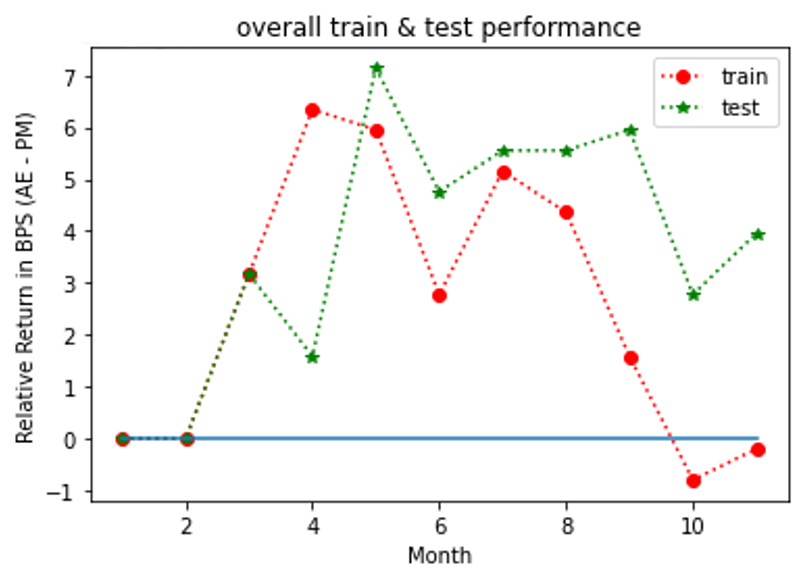}
	\caption{G-learner: overall trading performance with growth funds benchmarked by Russell 3000}
	\label{glearner_acc_avg_rs3k_growth}
\end{figure}

\begin{figure}[ht!]
	\centering
	\includegraphics[width=0.95\linewidth]{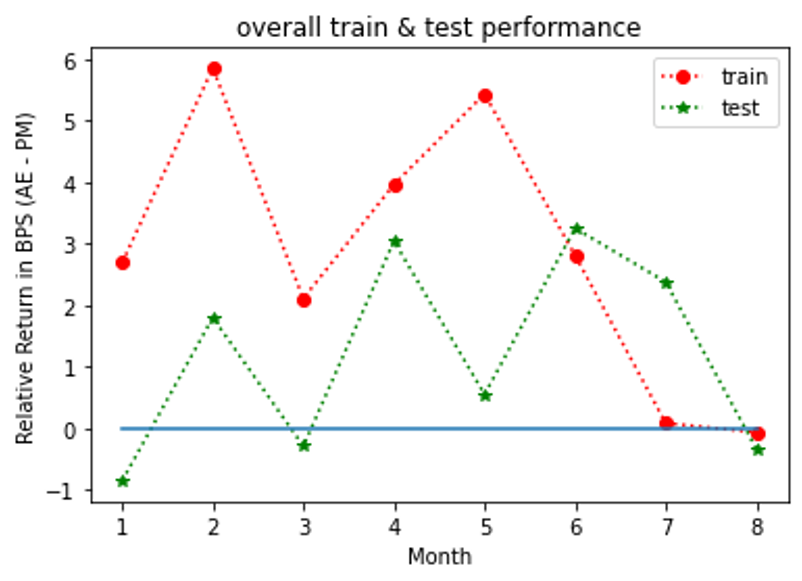}
	\caption{G-learner: overall trading performance with value funds benchmarked by Russell 3000}
	\label{glearner_acc_avg_rs3k_value}
\end{figure}

Figures \ref{glearner_acc_avg_sp500}, \ref{glearner_acc_avg_rs3k_growth}, \ref{glearner_acc_avg_rs3k_value} show the outperformance of our RL policy from PM's trading strategies (i.e, $AE - PM$) for all three fund groups for both the training and test sets for forward period of 12 months, plotted as functions of their time arguments. We note that both curves for the training and test sets follow similar paths for at least seven months before they started to diverge (the curves for Russell 3000 value group diverged significantly after the eighth months' trades and thus were truncated for a better visualization effect).  In practice, we recommend to update the policy through a re-training once the training and test results 
start to diverge. In general, our results  suggest that our G-learner is able to generalize (i.e. perform well out-of-sample)   up to 6  months into the  future.

\begin{figure}[ht!]
	\centering
	\includegraphics[width=0.95\linewidth]{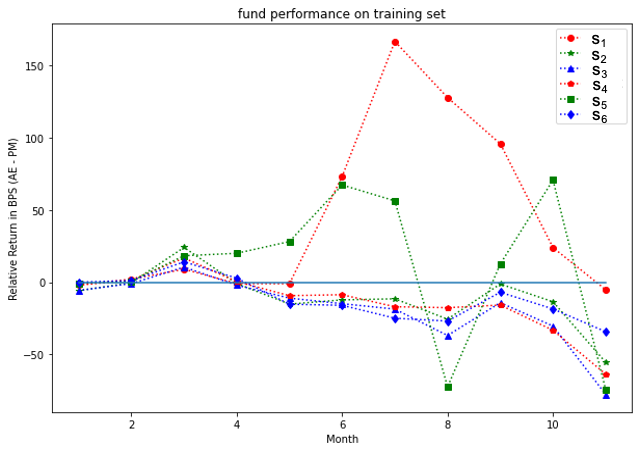}
	\caption{G-learner: trading performance of individual funds benchmarked by S\&P500 on the training set}
	\label{glearner_acc_each_sp500_train}
\end{figure}

\begin{figure}[ht!]
	\centering
	\includegraphics[width=0.95\linewidth]{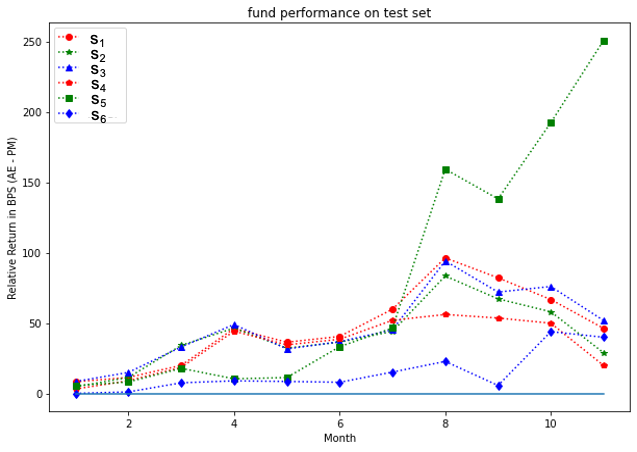}
	\caption{G-learner: trading performance of individual funds benchmarked by S\&P500 on the test set}
	\label{glearner_acc_each_sp500_test}
\end{figure}

\begin{figure}[ht!]
	\centering
	\includegraphics[width=0.95\linewidth]{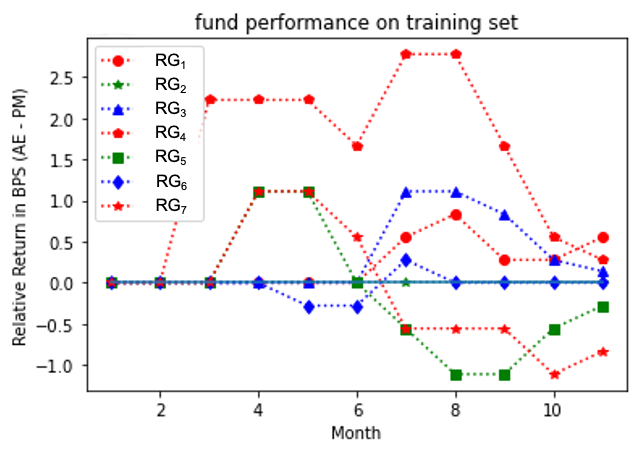}
	\caption{G-learner: trading performance of individual growth funds benchmarked by Russell 3000 on the training set}
	\label{glearner_acc_each_rs3k_growth_train}
\end{figure}

\begin{figure}[ht!]
	\centering
	\includegraphics[width=0.95\linewidth]{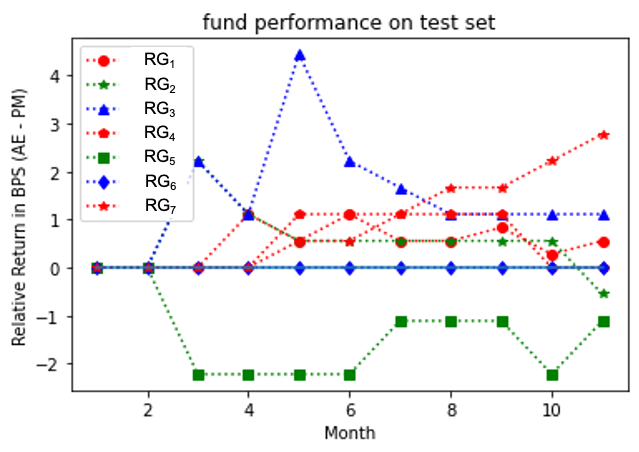}
	\caption{G-learner: trading performance of individual growth funds benchmarked by Russell 3000 on the test set}
	\label{glearner_acc_each_rs3k_growth_test}
\end{figure}

\begin{figure}[ht!]
	\centering
	\includegraphics[width=0.95\linewidth]{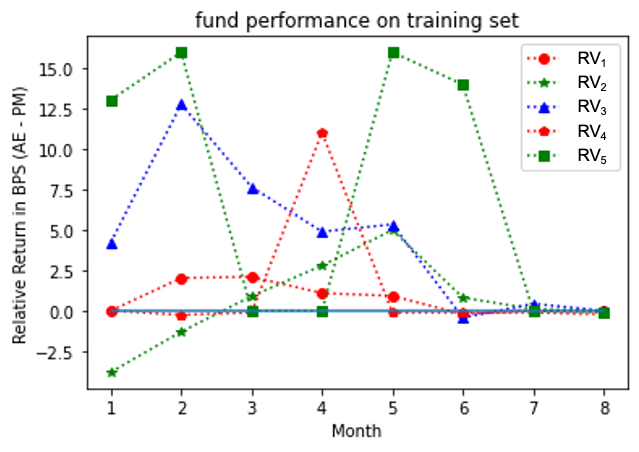}
	\caption{G-learner: trading performance of individual value funds benchmarked by Russell 3000 on the training set}
	\label{glearner_acc_each_rs3k_value_train}
\end{figure}

\begin{figure}[ht!]
	\centering
	\includegraphics[width=0.95\linewidth]{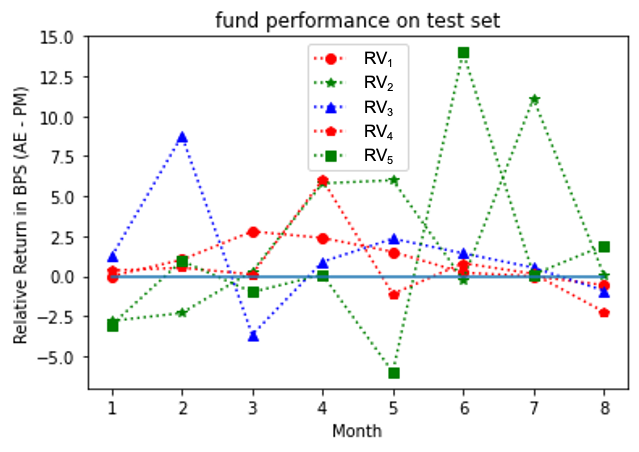}
	\caption{G-learner: trading performance of individual value funds benchmarked by Russell 3000 on the test set}
	\label{glearner_acc_each_rs3k_value_test}
\end{figure}

Analysis at the individual fund level for all three groups is presented in Figures \ref{glearner_acc_each_sp500_train}, \ref{glearner_acc_each_rs3k_growth_train}, \ref{glearner_acc_each_rs3k_value_train} for the training set, and Figures \ref{glearner_acc_each_sp500_test}, \ref{glearner_acc_each_rs3k_growth_test}, \ref{glearner_acc_each_rs3k_value_test} for the test set. Our model outperformed most of PMs throughout the test period except funds $RG_5$ and $RV_5$. 
It is interesting to note that the test results outperformed the training ones for the S\&P 500 fund group. This may be due to a potential market regime drift (which is favorable to us, in the present case).  More generally, a detailed study would be desired to address the impact of potential market drift or a regime change on optimal  asset  allocation policies.  Such topics are  left here for a future research.

\section{Conclusion and Future Work}
\label{sect_Summary}
In this work, we presented a first practical two-step procedure that combines the human and artificial intelligence for optimization of asset allocation and investment portfolios. At the first step, we use inverse reinforcement learning (IRL), and more specifically the parametric T-REX algorithm, to infer the reward function that captures fund managers'  intent of achieving higher returns. At the second step, we 
use the direct reinforcement learning with the inferred reward function
to improve the investment policy. Our approach is modular, and allows one to use different IRL or RL models, if needed, instead of the particular combination of the parametric T-REX and G-learner that we used in this work. It can also use other ranking criteria, e.g. sorting by the Sharpe ratio or the Sortino ratio, instead of ranking trajectories by their total return as was done above.    

Using aggregation of equity positions at the sector level, we showed 
that our approach is able to outperform most of fund managers. 
Outputs of  the model can be used by individual fund managers by keeping their stock selection and re-weighting their portfolios across industrial sectors according to the recommendation from G-learner. A practical implementation of our method should involve checking the feasibility of recommended allocations and controlling for potential market impact effects and/or transaction costs resulting from such rebalancing. 

Finally, we note that while in this work we choose a particular method of dimensional reduction that aggregates all stocks at the industry level, this is not the only available choice. An alternative approach could be considered, where all stocks are aggregated based on their factor exposure. This is left here for a future work.

\bibliography{citation}
\bibliographystyle{icml2020}

\end{document}